\documentclass[10pt,twocolumn,letterpaper]{article}

\usepackage{cvpr}              

\usepackage{epsfig}
\usepackage{graphicx}
\usepackage{amsmath}
\usepackage{amssymb}
\usepackage{enumitem}
\usepackage[normalem]{ulem}
\usepackage[ruled]{algorithm2e}
\usepackage{multirow}
\usepackage{array}
\usepackage{ctable} 
\usepackage{makecell}
\usepackage[capposition=bottom]{floatrow}
\usepackage{comment}
\DeclareMathAlphabet{\mathcal}{OMS}{cmsy}{m}{n}
\usepackage{textcomp}
\usepackage{dblfloatfix}
\usepackage{caption}
\usepackage{subcaption}
\usepackage{colortbl,xcolor}
\usepackage{booktabs}
\usepackage{mathtools}
\usepackage{textpos}
\newcommand{\name}{\textsc{ObjectFolder} 2.0\xspace}
\newcommand{\namecorl}{\textsc{ObjectFolder} 1.0\xspace}

\definecolor{MyDarkBlue}{rgb}{0,0.08,1}
\definecolor{MyAqua}{rgb}{0,0.7,0.7}
\definecolor{MyDarkGreen}{rgb}{0.02,0.6,0.02}
\definecolor{MyDarkRed}{rgb}{0.8,0.02,0.02}
\definecolor{MyDarkOrange}{rgb}{0.40,0.2,0.02}
\definecolor{MyPurple}{RGB}{111,0,255}
\definecolor{MyRed}{rgb}{1.0,0.0,0.0}
\definecolor{MyGold}{rgb}{0.75,0.6,0.12}
\definecolor{MyDarkgray}{rgb}{0.66, 0.66, 0.66}
\definecolor{nicegreen}{rgb}{0.1, 0.6, 0.2}

\definecolor{customgray}{rgb}{0.9, 0.9, 0.9}
\newcolumntype{g}{>{\columncolor{customgray}}c}
\newcolumntype{z}{>{\columncolor{customgray}}l}
\newcolumntype{?}[1]{!{\vrule width #1}}
\renewcommand{\paragraph}[1]{{\vspace{2mm}\noindent\textbf{#1}\,\,}}

\usepackage[pagebackref,breaklinks,colorlinks]{hyperref}

\usepackage[capitalize]{cleveref}
\crefname{section}{Sec.}{Secs.}
\Crefname{section}{Section}{Sections}
\Crefname{table}{Table}{Tables}
\crefname{table}{Tab.}{Tabs.}

\def\cvprPaperID{2475} 
\def\confName{CVPR}
\def\confYear{2022}

\begin{document}

\title{\name: A Multisensory Object Dataset for Sim2Real Transfer}

\author{Ruohan Gao\textsuperscript{1}\thanks{indicates equal contribution.} \hspace{5mm} Zilin Si\textsuperscript{2}\footnotemark[1] \hspace{10mm}  Yen-Yu Chang\textsuperscript{1}\footnotemark[1] \hspace{10mm} Samuel Clarke\textsuperscript{1}\\ 
Jeannette Bohg\textsuperscript{1} \hspace{10mm} Li Fei-Fei\textsuperscript{1} \hspace{10mm} Wenzhen Yuan\textsuperscript{2} \hspace{10mm} Jiajun Wu\textsuperscript{1}\\
\textsuperscript{1}Stanford Univeristy \hspace{10mm} \textsuperscript{2}Carnegie Mellon University\\
}

\maketitle
\pagestyle{empty}
\thispagestyle{empty}

\begin{abstract}
Objects play a crucial role in our everyday activities. Though multisensory object-centric learning has shown great potential lately, the modeling of objects in prior work is rather unrealistic. \namecorl~is a recent dataset that introduces 100 virtualized objects with visual, acoustic, and tactile sensory data. However, the dataset is small in scale and the multisensory data is of limited quality, hampering generalization to real-world scenarios. We present \name, a large-scale, multisensory dataset of common household objects in the form of implicit neural representations that significantly enhances \namecorl in three aspects. First, our dataset is 10 times larger in the amount of objects and orders of magnitude faster in rendering time. Second, we significantly improve the multisensory rendering quality for all three modalities. Third, we show that models learned from virtual objects in our dataset successfully transfer to their real-world counterparts in three challenging tasks: object scale estimation, contact localization, and shape reconstruction. \name~offers a new path and testbed for multisensory learning in computer vision and robotics. The dataset is available at \url{https://github.com/rhgao/ObjectFolder}.
\end{abstract}
\begin{textblock*}{\textwidth}(0cm,-17cm)
\centering
In Proceedings of the IEEE Conference on Computer Vision and Pattern Recognition (CVPR), 2022.%
\end{textblock*}
\vspace{-0.1in}
\section{Introduction}~\label{sec:intro}
\vspace{-0.0in}

Our everyday activities involve perception and manipulation of a wide variety of \emph{objects}. For example, we begin the morning by first turning off the alarm clock on the nightstand, slowly waking up. Then we may put some bread on a plate and enjoy our breakfast with a fork and knife to kick off the day. Each of these objects has very different physical properties---3D shapes, appearance, and material types, leading to their distinctive sensory modes: the alarm clock looks round and glossy, the plate clinks when struck with the fork, the knife feels sharp when touched on the blade. 

However, prior work on modeling real-world objects is rather limited and unrealistic. In computer vision, objects are often modeled in 2D with the focus of identifying and locating them in static images~\cite{deng2009imagenet,lin2014microsoft,girshick2014rich}. Prior works on shape modeling build 3D CAD models of objects~\cite{wu20153d,chang2015shapenet}, but they tend to focus purely on geometry, and the visual textures of the objects are of low-quality. Moreover, most works lack the full spectrum of physical object properties and focus on a single modality, mostly vision.

\begin{figure}
    \center
    \vspace{-0.05in}
    \includegraphics[scale=0.38]{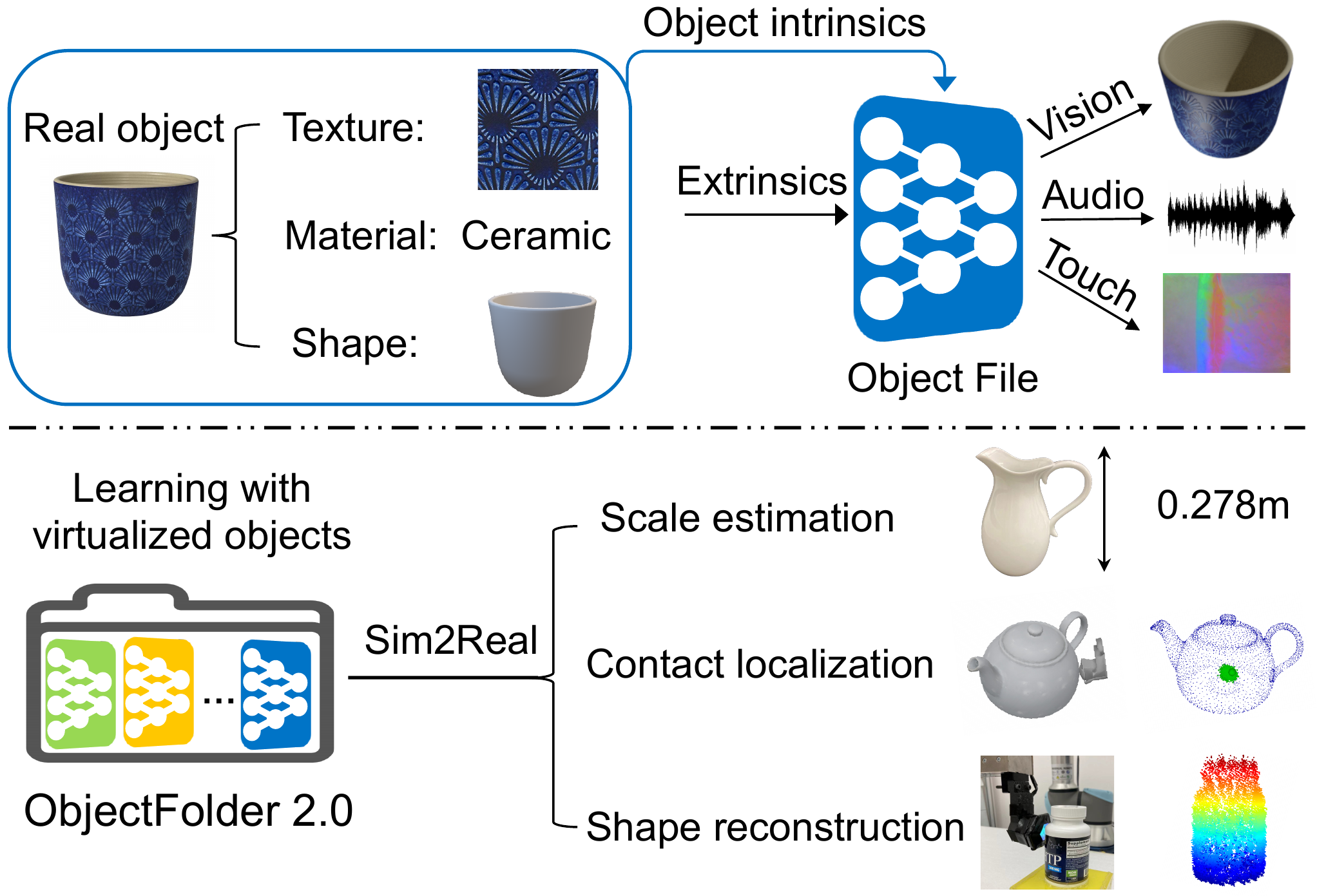}
    \caption{\name~contains 1,000 implicitly represented objects each containing the complete multisensory profile of a real object. We virtualize each object by encoding its intrinsics (texture, material type, and 3D shape) with an \emph{Object File} implicit neural representation. Then we can render its visual appearance, impact sound, and tactile readings based on any extrinsic parameters. We successfully transfer the models learned from our virtualized objects to three challenging tasks on their real-world counterparts. This opens a new path for multisensory learning in computer vision and robotics, where \name~serves as a rich and realistic object repository for training real-world models.}
    \vspace{-0.05in}
    \label{fig:concept}
\end{figure}

Our goal is to build a large dataset of realistic and multisensory 3D object models such that learning with these virtualized objects can generalize to their real-world counterparts. As shown in Fig.~\ref{fig:concept}, we leverage existing high-quality scans of real-world objects and extract their physical properties including visual textures, material types, and 3D shapes. Then we simulate the visual, acoustic, and tactile data for each object based on their object intrinsics, and use an implicit neural representation network---\emph{Object File}---to encode the simulated multisensory data. If the sensory data is realistic enough, models learned with these virtualized objects can then be transferred to real-world tasks involving these objects.

To this end, we introduce \name, a large dataset of implicitly represented multisensory replicas of real-world objects. It contains 1,000 high-quality 3D objects collected from online repositories~\cite{googleScannedObjects,collins2021abo,3dmodelhaven,calli2015ycb}. Compared with \namecorl\footnote{Throughout, we refer the \namecorl~\cite{gao2021ObjectFolder} dataset as 1.0 and our dataset as 2.0 for convenience.} that is slow in rendering and of limited quality in multisensory simulation, we improve the acoustic and tactile simulation pipelines to render more realistic multisensory data. Furthermore, we propose a new implicit neural representation network that renders visual, acoustic, and tactile sensory data all in real-time with state-of-the-art rendering quality. We successfully transfer models learned on our virtualized objects to three challenging real-world tasks, including object scale estimation, contact localization, and shape reconstruction. 

\name enables many applications, including 1) multisensory learning with vision, audio, and touch; 2) robot grasping of diverse real objects on various robotic platforms; and 3) applications that need on-the-fly multisensory data such as on-policy reinforcement learning.

In summary, our main contributions are as follows: First, we introduce a new large multisensory dataset of 3D objects in the form of implicit neural representations, which is 10 times larger in scale compared to existing work. We significantly improve the multisensory rendering quality for vision, audio, and touch, while being orders of magnitude faster in rendering time. Second, we show that learning with our virtualized objects can successfully transfer to a series of real-world tasks, offering a new path and testbed for multisensory learning for computer vision and robotics.

\section{Related Work}~\label{sec:related}
\vspace{-0.1in}

\paragraph{Object datasets.} Objects are modeled in different ways across different datasets. Image datasets such as ImageNet~\cite{deng2009imagenet} and MS COCO~\cite{lin2014microsoft} model objects in 2D. Datasets of synthetic 3D CAD models such as ModelNet~\cite{wu20153d} and ShapeNet~\cite{chang2015shapenet} focus on the geometry of objects without modeling their realistic visual textures. Pix3D~\cite{pix3d}, IKEA Objects~\cite{lim2013parsing}, and Object3D~\cite{xiang2016objectnet3d} align 3D CAD models to objects in real images, but they are either limited in size or make unignorable approximations in the 2D-3D alignment. BigBIRD~\cite{singhbigbird} and YCB~\cite{calli2015ycb} directly model real-world objects but only for a small number of object instances. ABO~\cite{collins2021abo} was recently introduced, containing 3D models for over 8K objects of real household objects, but it focuses only on the visual modality, similar to the other datasets above.

Alternatively, \name~contains 1,000 3D objects in the form of implicit neural representations, each of which encodes realistic visual, acoustic, and tactile sensory data for the corresponding object. Compared to \namecorl
~\cite{gao2021ObjectFolder}, our dataset is not only 10 times larger in the amount of objects, but also we significantly improve the quality of the multisensory data while being 100 times faster in rendering time. Furthermore, while \namecorl only performs tasks in simulation, we show that learning with our virtualized objects generalizes to the objects' real-world counterparts. 

\vspace{-0.05in}
\paragraph{Implicit neural representations.} Coordinate-based multi-layer perceptrons (MLPs) have attracted much attention lately and have been used as a new way to parameterize different types of natural signals. They are used to learn priors over shapes~\cite{park2019deepsdf,mescheder2019occupancy,chen2019learning}; represent the appearance of static scenes~\cite{mildenhall2020nerf, sitzmann2019scene}, dynamic scenes~\cite{Niemeyer2019ICCV,park2021nerfies}, or individual objects~\cite{guo2020object,Niemeyer2020GIRAFFE}; and even encode other non-visual modalities such as wavefields, sounds, and tactile signals~\cite{gao2021ObjectFolder,sitzmann2020implicit}. 

We also use MLPs to encode object-centric visual, acoustic, and tactile data similar to~\cite{gao2021ObjectFolder}, but our new object-centric implicit neural representations encode the intrinsics of objects more realistically and flexibly. Furthermore, inspired by recent techniques~\cite{liu2020neural,yu2021plenoctrees,garbin2021fastnerf,reiser2021kilonerf,hedman2021snerg,neff2021donerf,lindell2021autoint} on speeding up neural volume rendering~\cite{kajiya1984ray}, we largely reduce the rendering time of visual appearance, making inference of all sensory modalities real-time. 

\vspace{-0.05in}
\paragraph{Multisensory learning.} A growing body of work leverages other sensory modalities as learning signals in addition to vision, with audio and touch being the most popular. For audio-visual learning, inspiring recent work integrates sound and vision for a series of interesting tasks, including self-supervised representation learning~\cite{owens2016ambient,owens2018audio,Korbar2018cotraining}, audio-visual source separation~\cite{gao2018objectSounds,zhao2018sound,gao2019co,gan2020music}, sound localization in video frames~\cite{Senocak_2018_CVPR,tian2018audio}, visually-guided audio generation~\cite{gao2019visualsound,morgadoNIPS18}, and action recognition~\cite{wu2016multi,gao2020listen}. For visuo-tactile learning, the two sensory modalities are used for cross-modal prediction~\cite{li2019connecting} and representation learning~\cite{pinto2016curious,lee2019making}. Touch is also used to augment vision for 3D shape reconstruction~\cite{smith20203d,suresh2021efficient}, robotic grasping~\cite{calandra2017feeling,calandra2018more}, and object contact localization~\cite{luo2015localizing}. Earlier work on modeling multisensory physical behavior of 3D objects~\cite{pai2001scanning} proposes a system to directly measure contact textures and sounds, but mainly for the purpose of better modeling virtual object interaction and creating animations.

\name~is a potential testbed for various multisensory learning tasks involving all three modalities. Different from the works above, instead of learning with certain sensory modalities for a particular task, our goal is to introduce a dataset of implicitly represented objects with realistic visual, acoustic, and tactile sensory data, making multisensory learning easily accessible to the computer vision and robotics community.

\section{A Large Repository of Diverse Objects}~\label{sec:dataset}
\vspace{-0.15in}

\name~contains 1,000 3D objects in the form of implicit neural representations. Among the 1,000 objects, we use all 100 objects from \namecorl~\cite{gao2021ObjectFolder}, which consists of high quality 3D objects from 3D Model Haven~\cite{3dmodelhaven}, YCB~\cite{calli2015ycb}, and Google Scanned Objects~\cite{googleScannedObjects}. The recently introduced ABO dataset~\cite{collins2021abo} is another rich repository of real-world 3D objects, containing about 8K object models with high-quality 3D meshes, which come from Amazon.com product listings. For each object, we obtain metadata such as category, material, color, and dimensions on the real product's publicly available webpage. We filter the dataset by material type and only keep objects of the following materials: ceramic, glass, wood, plastic, iron, polycarbonate, and steel. We visually inspect each object's product images to make sure the metadata is correct and keep the object if its material property is approximately homogeneous. These steps ensure that the selected objects are acoustically simulatable as will be described in Sec.~\ref{sec:audio}. In the end, we obtain 855 objects from the ABO dataset. Additionally, we obtain 45 objects of polycarbonate material type from Google Scanned Objects.

\begin{figure}[t]
    \center
    \includegraphics[scale=0.33]{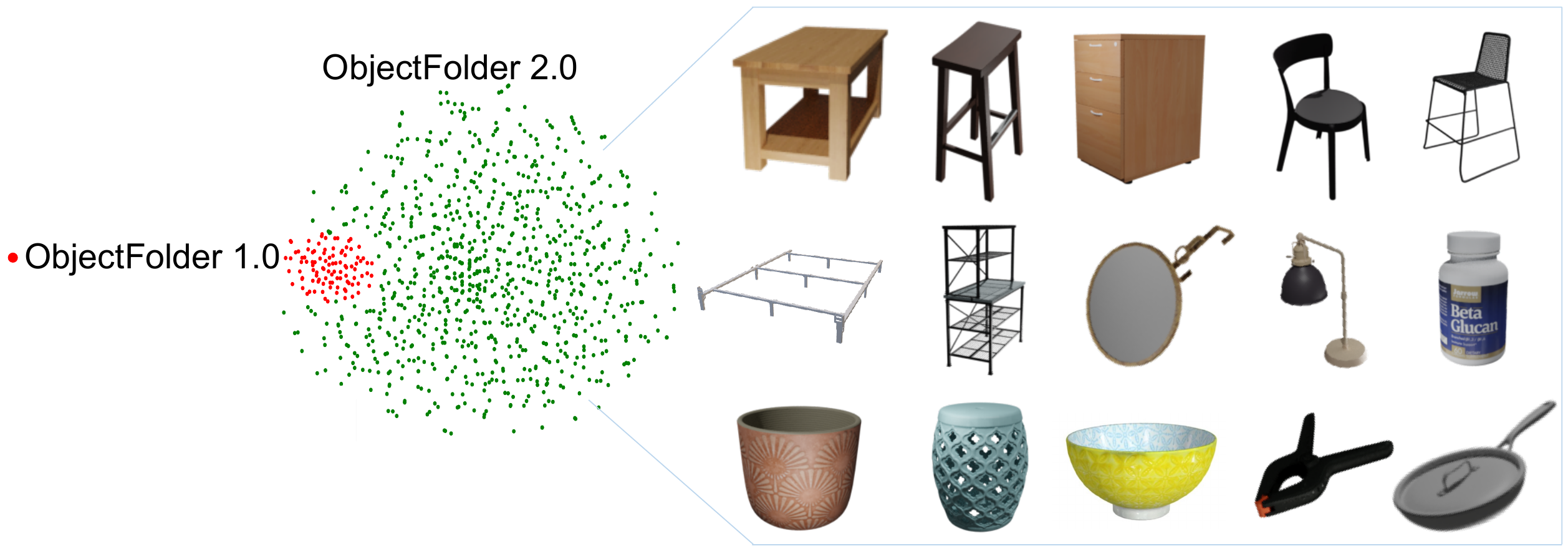}
    \caption{Example objects in \name. Each dot on the left represents an object in our dataset with red dots representing objects from \namecorl. 
    }
    \label{fig:dataset_example}
    \vspace{-0.1in}
\end{figure}

Fig.~\ref{fig:dataset_example} shows some example objects in our dataset.\footnote{Note that the original object meshes we use in our dataset all come from prior datasets~\cite{calli2015ycb,googleScannedObjects,collins2021abo}, and our contribution is a pipeline to create multisensory object assets based on these mesh models.} \name~is an order of magnitude larger than \namecorl and contains common household items of diverse categories including wood desks, ceramic bowls, plastic toys, steel forks, glass mirrors, etc.

\section{Improved Multisensory Simulation and Implicit Representations}~\label{sec:approach}
\vspace{-0.15in}

We propose a new simulation pipeline to obtain the multisensory data based on the objects' physical properties. Each object is represented by an \emph{Object File}, which is an implicit neural representation network that encodes the complete multisensory profile of the object. See Fig.~\ref{fig:concept}. Implicit representations have many advantages compared to conventional signal representations, which are usually discrete. We can parameterize each sensory modality as a continuous function that maps from some extrinsic parameters (e.g., camera view point and lighting conditions for vision, impact strength for audio, gel deformation for touch) to the corresponding sensory signal at a certain location or condition. Implicit neural representations serve as an approximation to this continuous function via a neural network. This makes the memory required to store the original sensory data independent of those extrinsic parameters, allowing the implicit representations to be easily streamed to users. Furthermore, thanks to the continuous property of implicit neural representations, the sensory data can be sampled at arbitrary resolutions.

\begin{figure*}[htbp]
    \centering
    \begin{subfigure}[b]{0.31\textwidth}
        \includegraphics[width=\textwidth]{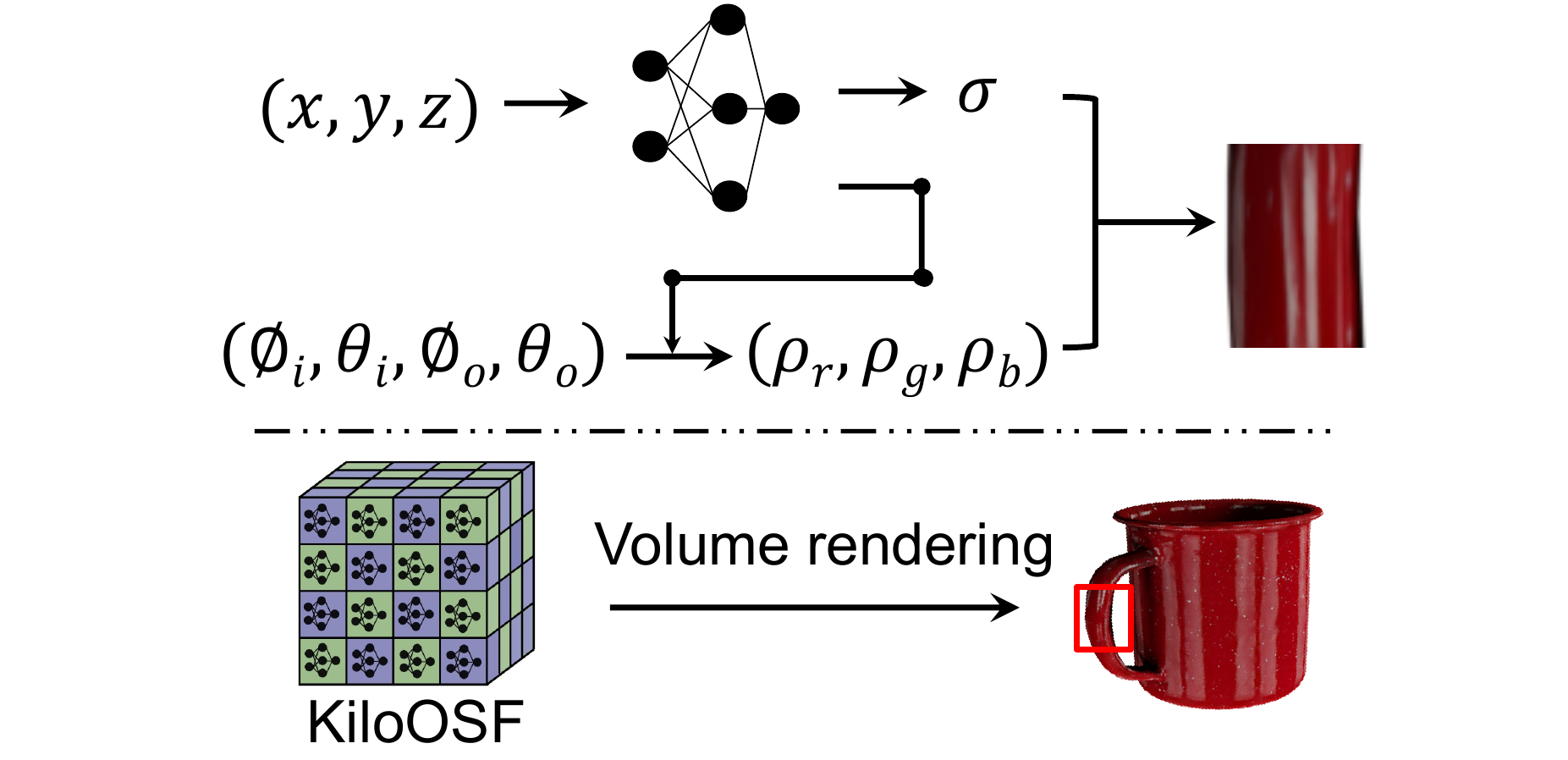}
        \caption{VisionNet}
        \label{fig:visionnet}
    \end{subfigure}
    \begin{subfigure}[b]{0.33\textwidth}
        \includegraphics[width=\textwidth]{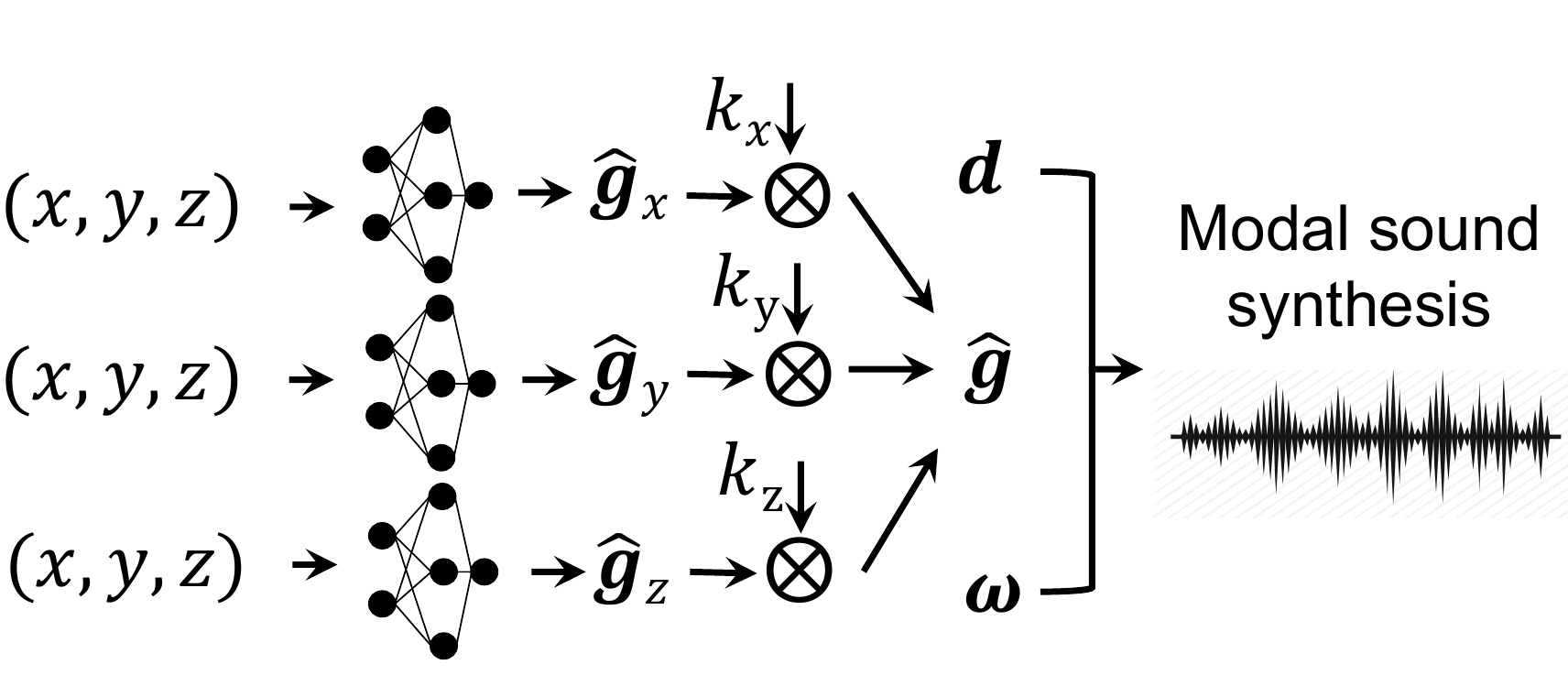}
        \caption{AudioNet}
        \label{fig:audionet}
    \end{subfigure}
    \begin{subfigure}[b]{0.35\textwidth}
        \includegraphics[width=\textwidth]{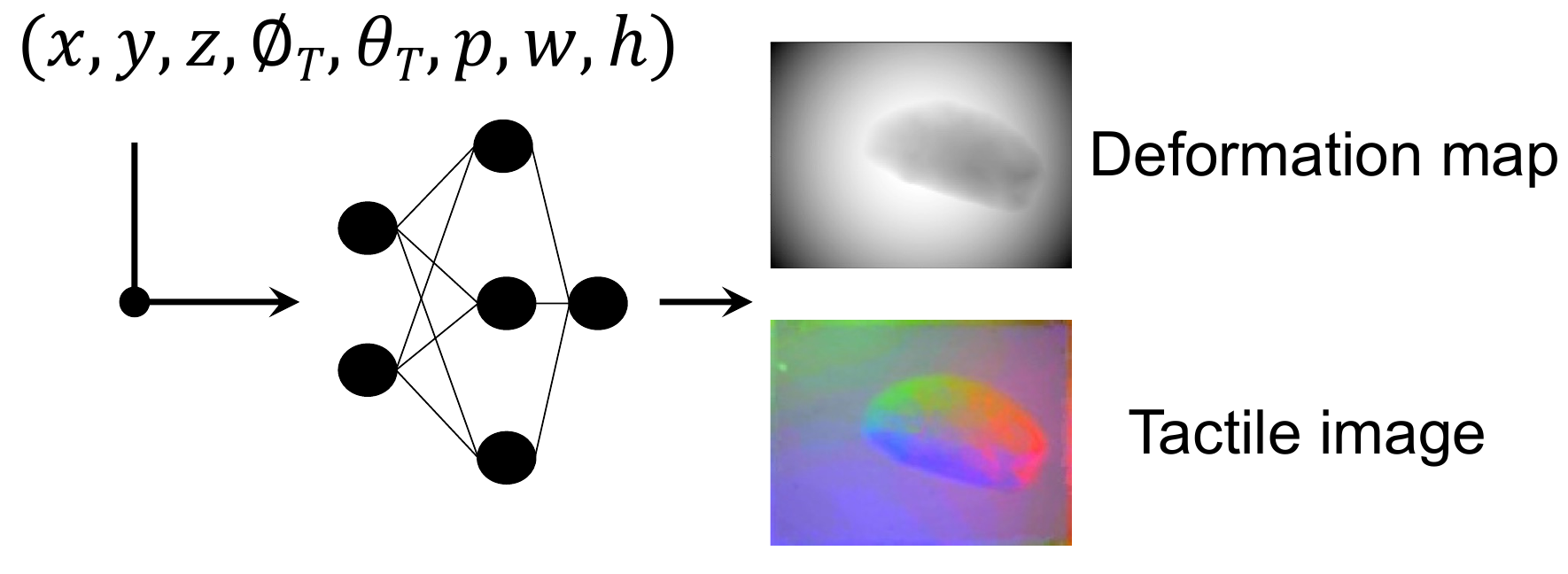}
        \caption{TouchNet}
        \label{fig:touchnet}
    \end{subfigure}
    \caption{Each \emph{Object File} implicit neural representation network contains three sub-networks: VisionNet, AudioNet, and TouchNet. Compared with \namecorl, we greatly accelerate VisionNet inference by representing each object with thousands of individual MLPs; for AudioNet, we only predict the parts of the signal that are location-dependent instead of directly predicting the audio spectrograms, which significantly improves the rendering quality and also accelerates inference; our new TouchNet can render tactile readings of varied rotation angles and gel deformations, whereas only a single tactile image can be rendered per vertex in 1.0.
    }
    \vspace{-0.05in}

    \label{fig:network}
\end{figure*}

Each \emph{Object File} has three sub-networks: VisionNet, AudioNet, and TouchNet (see Fig.~\ref{fig:network}). In the following, we introduce the details of how we simulate the three modalities and how we use multi-layer perceptrons (MLPs) to encode the data. 

\subsection{Vision}

\vspace{-5pt}\paragraph{Background.} Recent work~\cite{guo2020object} proposes to represent the appearance of each object by a neural network $F_v$ that models the object-centric neural scattering function (OSF).  $F_v$ takes as input a 3D location $\mathbf{x} = (x, y, z)$ in the object coordinate frame and the lighting condition at that location $(\omega_i, \omega_o)$, where $\omega_i = (\phi_i, \theta_i)$ and $\omega_o = (\phi_o, \theta_o)$ denote the incoming and outgoing light directions, respectively. The output is the volume density $\sigma$ and fraction of the incoming light that is scattered in the outgoing direction $\mathbf{\rho} = (\rho_r, \rho_g, \rho_b)$.
The amount of light scattered at a point $\mathbf{x}$ can be obtained as follows:
\vspace{-0.05in}
\begin{equation}
    L_s (\mathbf{x},\mathbf{\omega_o}) = \int_{S} L(\mathbf{x}, \omega_i)f_{\rho}(\mathbf{x}, \mathbf{\omega_i}, \mathbf{\omega_o})d\mathbf{\omega_i},
\end{equation}
where $S$ is a unit sphere, $L(\mathbf{x}, \omega_i)$ denotes the amount of light scattered at point $\mathbf{x}$ along direction $\omega_i$, and $f_{\rho}$ evaluates the fraction of light incoming from direction $\mathbf{\omega_i}$ at the point that scatters out in direction $\mathbf{\omega_o}$. 

Classic volume rendering~\cite{kajiya1984ray} is then used to render the color of any ray passing through the object. To render a single image pixel, a ray is cast from the camera's eye through the pixel's center. We denote the direction of the camera ray as $\mathbf{r}(t) = \mathbf{x}_0 + t \mathbf{\omega_o}$. A number of points $\mathbf{x}_1, \mathbf{x_2}, \ldots, \mathbf{x}_K$ are sampled along the ray. The final expected color $C(\mathbf{r})$ of camera ray $\mathbf{r}(t)$ can be obtained by $\alpha$-blending the list of $K$ color values ($L_s(\mathbf{x}_1, \mathbf{\omega_o}), L_s(\mathbf{x}_2, \mathbf{\omega_o}), \ldots, L_s(\mathbf{x}_K, \mathbf{\omega_o})$) with the following equation:
\vspace{-0.05in}
\begin{equation}
    C(\mathbf{r}) = \sum_{i=1}^{K} T_i (1 - \text{exp}(-\sigma_i \delta_i)) L_s(\mathbf{x}_i, \mathbf{\omega_o}),
\end{equation}
where $T_i = \text{exp}(-\sum_{j=1}^{i-1} \sigma_j \delta_j)$ denotes the accumulated transmittance along the ray, and $\delta_i = \|\mathbf{x}_{i+1} - \mathbf{x}_i \| $ denotes the distance between two adjacent sample points.

\paragraph{KiloOSF.} The above process has to be repeated for every pixel to render an image. Due to the large number of required forward passes through $F_v$, this makes it very time-consuming even on high-end consumer GPUs.

Inspired by many recent works on speeding up neural rendering~\cite{yu2021plenoctrees,garbin2021fastnerf,reiser2021kilonerf,hedman2021snerg,neff2021donerf}, we build upon KiloNeRF~\cite{reiser2021kilonerf} and introduce KiloOSF as our VisionNet. Instead of using a single MLP to represent the entire scene, KiloNeRF represents the static scene with a large number of independent and small MLPs. Each individual MLP is assigned a small portion of the scene, making each small network sufficient for photo-realistic rendering.

Similarly, we subdivide each \emph{object} into a uniform grid of resolution $\mathbf{s} = (s_x, s_y, s_z)$ with each grid cell of 3D index $\mathbf{i} = (i_x, i_y, i_z)$. Then we define a mapping $m$ from position $\mathbf{x}$ to index $\mathbf{i}$ through the following spatial binning:
\vspace{-0.05in}
\begin{equation}
\vspace{-0.05in}
    m(\mathbf{x}) =  \lfloor (\mathbf{x} - \mathbf{b}_{\text{min}}) / ((\mathbf{b}_{\text{max}} - \mathbf{b}_{\text{min}}) / \mathbf{s}) \rfloor,
\end{equation}
where $\mathbf{b}_{\text{min}}$ and $\mathbf{b}_{\text{max}}$ are the respective minimum and maximum bounds of the axis aligned bounding box (AABB) enclosing the object. For each grid cell, a tiny MLP network with parameters $v(\mathbf{i})$ is used to represent the corresponding portion of the object. Then, the color and density values at a point $\mathbf{x}$ and direction $\mathbf{r}$ can be obtained by first determining the index $m(\mathbf{x})$ responsible for the grid cell that contains this point, then querying the respective tiny MLP:
\vspace{-0.05in}
\begin{equation}
\vspace{-0.05in}
    (\mathbf{c}, \sigma) = F_{v(m(\mathbf{x}))} (\mathbf{x}, \mathbf{r}).
\end{equation}

Following KiloNeRF~\cite{reiser2021kilonerf}, we use a ``training with distillation" strategy to avoid artifacts in rendering. We first train an ordinary OSF~\cite{guo2020object} model for each object and then distill the knowledge of the teacher model into the KiloOSF model. We also use empty space skipping and early ray termination to increase rendering efficiency. See~\cite{reiser2021kilonerf} for details. Compared with \namecorl, our new VisionNet design significantly accelerates the rendering process at inference time 60 times (see Table~\ref{Table:rendering_time_comparison}) while simultaneously achieving better visual rendering quality.

\subsection{Audio}
\label{sec:audio}

\vspace{-5pt}\paragraph{Background.} Linear modal analysis is a standard way to perform physics-based 3D modal sound synthesis~\cite{ren2013example,jin2020deep,wang2019kleinpat}. A 3D linear elastic dynamic system can be modeled with the following linear deformation equation:
\vspace{-0.05in}
\begin{equation}
\vspace{-0.05in}
    \mathbf{M}\ddot{\mathbf{x}} + \mathbf{C}\dot{\mathbf{x}} + \mathbf{K}\mathbf{x} = \mathbf{f},
\end{equation}
where $\mathbf{x}$ denotes the nodal displacement, and $\mathbf{M}$, $\mathbf{C} = \alpha\mathbf{M} + \beta\mathbf{K}$, $\mathbf{K}$ represent the mass, Rayleigh damping, and stiffness matrices, respectively.\footnote{The values of these matrices depend on the object's scale and material. See Supp. for the mapping from material type to material parameters.} $\mathbf{f}$ represents the external nodal force applied to the object that stimulates the vibration. Through generalized eigenvalue decomposition $\mathbf{KU} = \mathbf{\Lambda MU}$, the above equation can be reformulated into the following form:
\begin{equation}
    \ddot{\mathbf{q}} + (\alpha\mathbf{I} + \beta\mathbf{\Lambda})\dot{\mathbf{q}} + \mathbf{\Lambda q} = \mathbf{U^Tf},
\vspace{-0.05in}
\end{equation}
where $\mathbf{\Lambda}$ is a diagonal matrix, and $\mathbf{q}$ satisfies $\mathbf{x} = \mathbf{Uq}$. The solution to the above equation is $N$ damped sinusoidal waves, each representing a mode signal. The $i_\text{th}$ mode is
\begin{equation}
\vspace{-0.05in}
    q_i = g_i e^{-d_i t} \sin(2\pi \omega_i t),~i = \{1,2,\ldots,N\}
\end{equation}
where $\omega_i$, $d_i$, and $g_i$ represent the damped natural frequencies, damping coefficients, and gains of the modes signals, respectively. Note that the gains $g_i$ of each mode are specific to the contact force and the contact location on the object, while the frequencies $\omega_i$ and damping coefficients $d_i$ of each mode are intrinsic parameters of the object.

\paragraph{AudioNet.} We convert the surface mesh of each object into a volumetric quadratic tedrahedral mesh using a sequential approach designed for object meshes from the wild~\cite{hu2018tetrahedral}, then use Finite Element Methods (FEM)~\cite{hughes2012finite} on the resultant tetrahedral mesh with second-order elements in Abaqus~\cite{abaqus2021dassault} to perform the modal analysis process described above. We simulate the vibration modes from contacting each vertex on the tedrahedral mesh with unit force in each axis direction. Then, we train an MLP that takes the vertex coordinate of the tedrahedral mesh as input and predicts the vector of gains of each mode for that vertex when contacted by unit force for each axis direction. 

At inference time, an object's impulse response can be predicted by first using the network to predict the gains $g_i$ of each mode, then constructing the response by summing the exponentially decaying sinusoids parameterized by the gains $\hat{g}_i$ predicted from the network, along with the frequencies $\omega_i$ and dampings $d_i$ obtained from modal analysis. We decompose the external force $\mathbf{f}$ at a vertex into a linear combination of unit forces along the three orthogonal axis directions: $\mathbf{f} = k_x \mathbf{f}_x + k_y \mathbf{f}_y + k_z \mathbf{f}_z$. The predicted gains $\mathbf{\hat{g}}$ excited by $\mathbf{f}$ can be obtained as follows: $\mathbf{\hat{g}} = k_x \mathbf{\hat{g}}_x + k_y\mathbf{\hat{g}}_y + k_z\mathbf{\hat{g}}_z$, where $\mathbf{\hat{g}}_x, \mathbf{\hat{g}}_y, \mathbf{\hat{g}}_z$ denote the the respective gains obtained from the three branches of AudioNet. Finally, combining the frequencies $\boldsymbol{\omega}$ and damping coefficients $\mathbf{d}$, we synthesize the audio waveform:
\vspace{-0.05in}
\begin{equation}
    S(t) = \sum_{i=1}^{N} \hat{g}_i e^{-d_i t} \sin(2\pi \omega_i t),
\vspace{-0.05in}
\end{equation}
where $\hat{g}_i$, $d_i$, and $\omega_i$ represent elements of $\mathbf{\hat{g}}$, $\mathbf{d}$, and $\boldsymbol{\omega}$, respectively.

\begin{figure}[t]
    \center
    \includegraphics[scale=0.395]{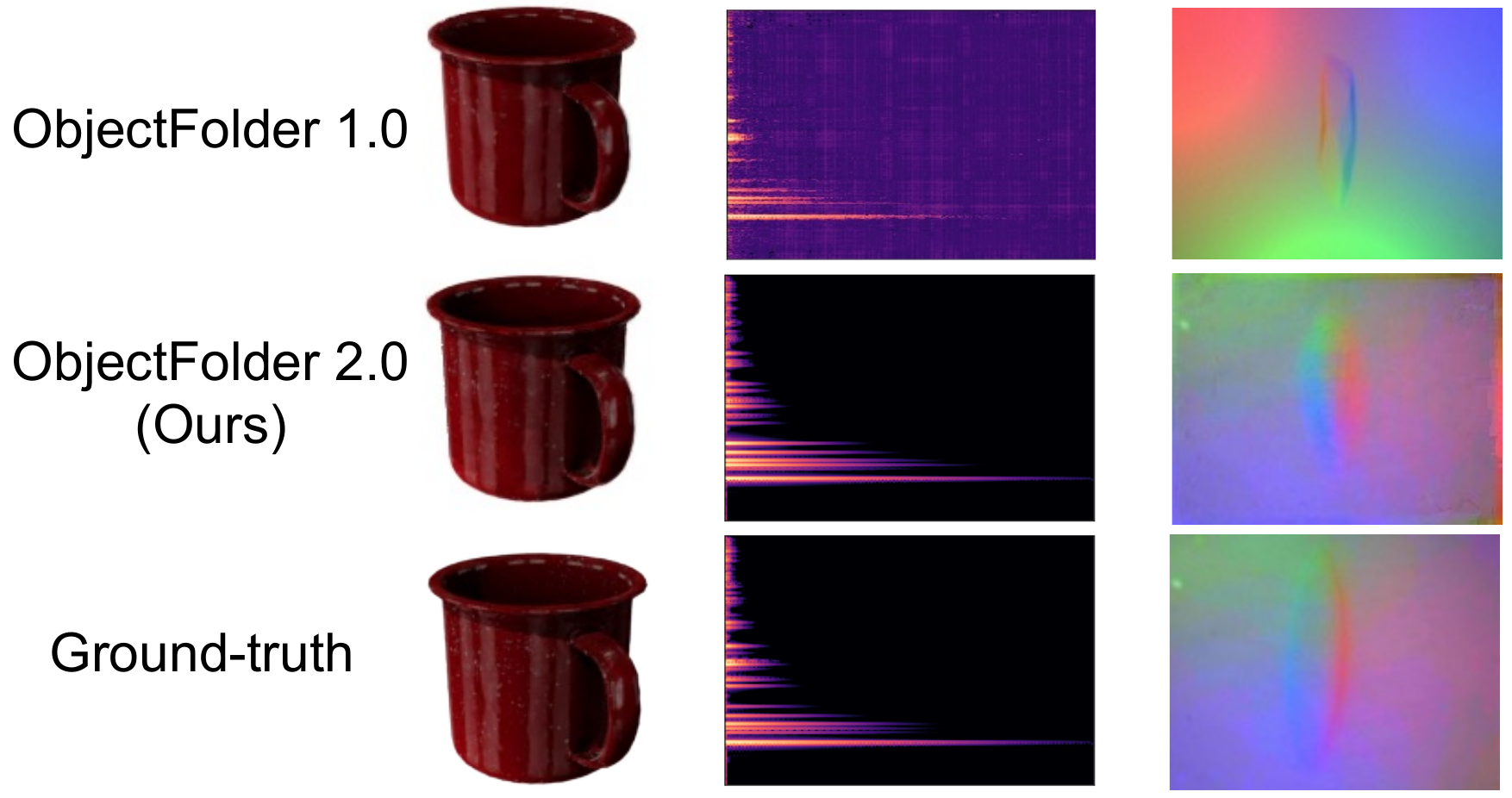}
    \caption{Comparing the visual, acoustic, and tactile data rendered from \namecorl, \name (Ours), and the corresponding ground-truth simulations for the YCB mug. See Supp. for more examples.
    }
    \label{fig:object_file_gt_comparison}
        \vspace{-0.1in}
\end{figure}

As opposed to using a volumetric hexahedron mesh for modal analysis as in \namecorl, the higher-order tetrahedral meshes we use for modal analysis capture finer features and surface curvature as well as more precise elastic deformations, at the same representation size. Thus it can more accurately model the acoustic properties of the objects~\cite{he2012fem,schneider2019large,bharaj2015computational}. 
Moreover, the AudioNet in 1.0 directly predicts a complex audio spectrogram, which is of much higher dimension and is limited to a fixed resolution and temporal length. We instead only predict the parts of the signal that are location-dependent, and then analytically obtain the remainder of the modes signal. This significantly improves the quality of audio rendering with our new implicit representation network. See  Table~\ref{Table:rendering_quality} and Fig.~\ref{fig:object_file_gt_comparison} for a comparison.

\begin{table}
{\resizebox{0.99\linewidth}{!}{
\small
\begin{tabular}{lcccc}
\toprule
     & Vision & Audio & Touch & Total \\ 
\midrule
     \namecorl~\cite{gao2021ObjectFolder}  & 3.699  & 0.420 & 0.010 & 4.129 \\ 
     \name (Ours) & 0.062 & 0.035 & 0.014 & 0.111   \\
\bottomrule
\end{tabular}
}}
\caption{Time comparison for rendering one observation sample for each modality, in seconds.}
\label{Table:rendering_time_comparison}
\vspace{-0.1in}
\end{table}

\vspace{-0.05in}
\subsection{Touch}

\vspace{-5pt}\paragraph{Background.} We use the geometric measurement from a GelSight tactile sensor~\cite{yuan2017gelsight,dong2017improved} as the tactile reading. GelSight is a vision-based tactile sensor that interacts the object with an elastomer and measures the geometry of the contact surface with an embedded camera. It has a very high spatial resolution of up to 25 micrometers and can potentially be used to synthesize readings from other tactile sensors~\cite{lambeta2020digit,padmanabha2020omnitact}. To simulate tactile sensing with GelSight, we need to simulate both the deformation of the contact and the optical response to the deformation. For our tactile simulation, we aim to achieve the following three goals: 1) Being flexible to render tactile readings for touches of varied location, orientation, and pressing depth; 2) Being fast to efficiently render data for training TouchNet; 3) Being realistic to generalize to real-world touch sensors.

\begin{table*}[ht]
{\resizebox{0.955\linewidth}{!}{
\small
\begin{tabular}{lcccccc}
\toprule
\multirow{2}{*}{} & \multicolumn{2}{c}{Vision} & \multicolumn{2}{c}{Audio} & \multicolumn{2}{c}{Touch} \\ 
\cmidrule(lr){2-3}\cmidrule(lr){4-5}\cmidrule(lr){6-7}
                  & PSNR~$\uparrow$    & SSIM~$\uparrow$     & STFT Distance ($\times10^{-5})$~$\downarrow$  & ENV Distance ($\times10^{-4})$~$\downarrow$ & PSNR~$\uparrow$  & SSIM~$\uparrow$    \\ 
\midrule
\namecorl~\cite{gao2021ObjectFolder}   &    35.7     &     0.97       &  4.94   &  7.65  & 27.9 & 0.64   \\ 
\name (Ours)        &  36.3      &       0.98   &  0.19       &  1.29   &  31.6  & 0.78 \\
\bottomrule
\end{tabular}
}}
\caption{Comparing with \namecorl on the multisensory data rendering quality. $\downarrow$ lower better, $\uparrow$ higher better.}
\vspace{-0.1in}
\label{Table:rendering_quality}
\end{table*}

\paragraph{TouchNet.} To achieve the three goals above, we adopt a two-stage approach to render realistic tactile signals. First, we simulate the contact deformation map, which is constructed from the object's shape in the contact area and the gelpad's shape in the non-contact area to represent the local shape at the point of contact. 
We simulate the sensor-object interaction with Pyrender~\cite{Pyrender} to render deformation maps using OpenGL~\cite{OpenGL} with GPU-acceleration, reaching 700 fps for data generation. 

We design TouchNet to encode the deformation maps from contacting each vertex on the object. We represent the tactile readings of each object as an 8D function whose input is a 3D location $\mathbf{x} = (x, y, z)$ in the object coordinate frame, a 3D unit contact orientation parametrized as $(\theta_T, \phi_T)$, gel penetration depth $p$, and the spatial location $(w, h)$ in the deformation map. The output is the per-pixel value of the deformation map for the contact. TouchNet models this continuous function as an MLP network $F_T: (x, y, z, \theta_T, \phi_T, p, w, h) \longrightarrow  d$ that maps each input 8D coordinate to its corresponding value in the deformation map. After rendering the deformation map, we utilize the state-of-the-art GelSight simulation framework--- Taxim~\cite{si2021taxim}, an example-based tactile simulation model that is calibrated with a real GelSight sensor, to render tactile RGB images from the deformation maps.

Compared to the TouchNet in~\namecorl, which can only render a single tactile image along the vertex normal direction per vertex, our new design of TouchNet can generate tactile outputs for rotation angles within $\pm15^{\circ}$ and pressing depth in the range of 0.5-2 mm. Furthermore, with the help of Taxim, the mapping from the deformation maps to the tactile optical outputs can be easily calibrated to different real vision-based tactile sensors, producing realistic tactile optical outputs that enable Sim2Real transfer.

\vspace{-0.3in}
\subsection{\textbf{\namecorl vs. \name}}

\name~significantly advances \namecorl in multisensory simulation and the design of implicit neural representations. Table~\ref{Table:rendering_time_comparison} shows the rendering time comparison. Our new network design is orders of magnitude faster compared to \namecorl, making rendering of all three sensory modalities real-time. The rendering quality is also greatly improved, especially for audio and touch as shown in the example of Fig.~\ref{fig:object_file_gt_comparison}. Our KiloOSF VisionNet renders images that match the ground-truth well while being $60\times$ faster than \namecorl. While directly predicting audio spectrograms cannot capture the details of the modes signal and leads to artifacts in the background, our AudioNet renders audio in a much more accurate manner. For touch, to make a fair comparison, we use the TACTO~\cite{wang2020tacto} simulation used in \namecorl and the tactile readings from real-world GelSight sensors as the ground truth instead. Our TouchNet output matches well with the real tactile readings.

Table~\ref{Table:rendering_quality} shows the quantitative comparisons. For visual and tactile rendering, we compare using standard metrics: peak signal-to-noise ratio (PSNR) and structural index similarity (SSIM) between the rendered image and the ground-truth image. For audio rendering, we report the STFT distance, which is the euclidean distance between the spectrograms of the ground-truth and the predicted modes signals, and the Envelope (ENV) Distance, which measures the Euclidean distance between the envelopes of the ground-truth and the predicted modes signals. For touch, because \namecorl uses the DIGIT~\cite{lambeta2020digit} tactile sensor, we compare with the real tactile images collected from a DIGIT sensor and a GelSight sensor for 1.0 and ours, respectively. Our TouchNet based on GelSight sensors has a smaller Sim2Real gap.

\begin{figure}
    \center
    \includegraphics[scale=0.27]{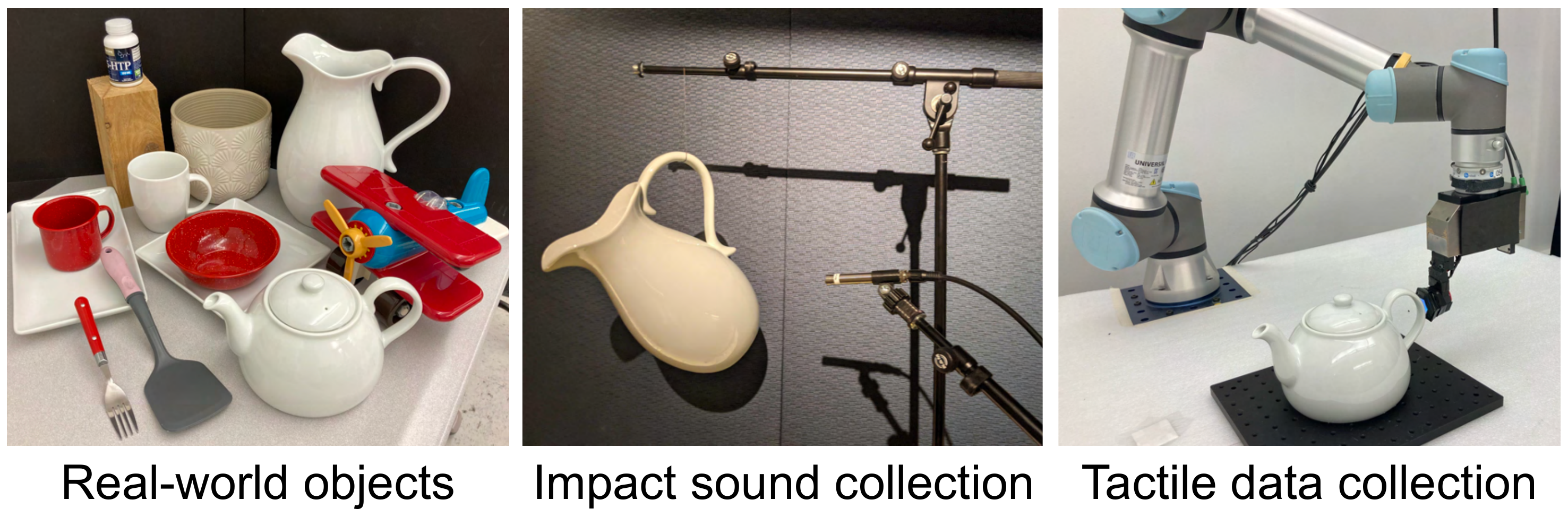}
    \caption{Illustration of real-world objects used in experiments and our hardware set-up for collecting real-world impact sounds and tactile data.}
    \label{fig:real_objects}
        \vspace{-0.1in}
\end{figure}

\vspace{-0.35in}
\section{Sim2Real Object Transfer}\label{sec:results}
\vspace{-0.05in}

The goal of building \name~is to enable generalization to real-world objects by learning with the virtual objects from our dataset. We demonstrate the utility of the dataset by evaluating on three tasks including object scale estimation, contact localization, and shape reconstruction. In each task, we transfer the models learned on \name~to real-world objects. See Fig.~\ref{fig:real_objects} for an illustration of the 13 objects used in our experiments, and the hardware set-up for collecting real impact sounds and GelSight tactile readings. 

\subsection{Object Scale Estimation}
\vspace{-0.05in}

All sensory modalities of objects are closely related to their scales. We want to demonstrate that learning with our virtualized objects can successfully transfer to scale estimation for a real object based on either its visual appearance, an impact sound, or a sequence of tactile readings. We train on the rendered multisensory data from our dataset, and test on 8 real objects from which we have collected real-world sensory data for all three modalities.

For vision and audio, we train ResNet-18~\cite{he2016deep} that takes either an RGB image of the object or the magnitude spectrogram of an impact sound as input to predict object scale\footnote{We define the scale of an object as the length of the longest side of the axis aligned bounding box (AABB) enclosing the object.}. From a single local tactile reading, it is almost impossible to predict the scale of the object. Therefore, we use a recurrent neural network to combine features from 10 consecutive touch readings for tactile-based scale prediction. See Supp. for details. 

Table~\ref{tab:scale_prediction} shows the results. ``Random'' denotes the baseline that randomly predicts a scale value within the same range as our models. We compare with models trained on sensory data from \namecorl. Both \namecorl and our dataset achieve high scale prediction accuracy on virtual objects. However, models trained on our multisensory data generalize much better to real-world objects, demonstrating the realism of our simulation and accurate encoding of our implicit representation networks. Among the three modalities, tactile data has the smallest Sim2Real gap compared to vision and audio.

\begin{table}
    \small
    \centering
    \begin{tabular}{@{}ll*{2}{c}}
    \toprule
    \multicolumn{1}{c}{} & & Virtual Objects    & Real Objects        \\ \midrule
 	& Random    & 14.5   & 14.5    \\ \midrule
 	\multirow{3}*{\rotatebox{90}{\footnotesize 1.0~\cite{gao2021ObjectFolder}}}
    & Vision & 0.80  & 7.41   \\
 	& Audio   & 0.57 & 6.85     \\ 
	& Touch   & 0.19  & 4.92     \\  \midrule
	\multirow{3}*{\rotatebox{90}{\footnotesize 2.0 (Ours)}}
	& Vision                  &     0.79       &  5.08                \\ 
	& Audio &   0.20       &    4.68         \\ 
	& Touch         &   0.45         &     3.51    \\  
   \bottomrule
    \end{tabular}
    \caption{Results on object scale prediction. We report the average difference between the predicted and the ground-truth scales of the objects in centimeters.}
    \label{tab:scale_prediction}
        \vspace{-0.1in}
\end{table}

\begin{table*}[ht]
\small
\begin{tabular}{lcccccccccccc}
\toprule
\multirow{2}{*}{Modalities} & \multicolumn{2}{c}{\includegraphics[height=.05\textwidth]{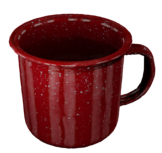}} & \multicolumn{2}{c}{ \includegraphics[height=.05\textwidth]{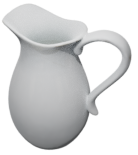}} &
\multicolumn{2}{c}{\includegraphics[height=.05\textwidth]{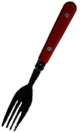}} & \multicolumn{2}{c}{\includegraphics[height=.05\textwidth]{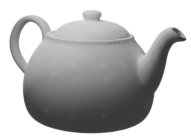}} &
\multicolumn{2}{c}{\includegraphics[height=.05\textwidth]{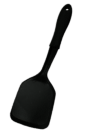}} & \multicolumn{2}{c}{\includegraphics[height=.05\textwidth]{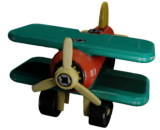}} 
 \\ 
\cmidrule(lr){2-3}\cmidrule(lr){4-5} \cmidrule(lr){6-7} \cmidrule(lr){8-9} \cmidrule(lr){10-11} \cmidrule(lr){12-13} 
                  & Sim   & Real    & Sim   & Real & Sim & Real  & Sim   & Real    & Sim   & Real & Sim & Real  \\ \midrule
Random   &    6.74     &     6.74       &  12.96  &  12.96  &  4.28   &  4.28  & 9.39 & 9.39 & 14.53 & 14.53 &    14.21     &     14.21    \\ 
Audio   &    1.88     &     1.79       &  0.26   &  1.16  &  0.65   &  4.67  & 0.23 & 1.04  & 0.14 & - &    0.74     &     -        \\ 
Touch        &  0.04      &   1.26     &     0.03       &  0.78   &     0.18       &  1.30   &  0.04  & 0.44 &  0.04  & 0.91 &  0.04      &   3.82  \\
Audio + Touch        &  0.02      &   0.59     &     0.04       &  0.36  &     0.09       &  0.51  &  0.04  & 0.63  &   0.23  & - &  0.30      &   -     \\ 
\bottomrule
\end{tabular}
\caption{Results on audio-tactile contact localization. We report the mean distance w.r.t. the ground-truth contact locations in centimeters.}
\label{Table:contact_localization}
\vspace{-0.1in}
\end{table*}

\vspace{-0.05in}
\subsection{Tactile-Audio Contact Localization}

When interacting with an object of known shape, accurately identifying the location where the interaction happens is of great practical interest. Touch gives local information about the contact location, and impact at varied surface locations produces different modal gains for the excited sound. We investigate the potential of using the impact sounds and/or the tactile readings associated with the interaction for contact localization.

We apply particle filtering~\cite{liu1998sequential} to localize the sequence of contact locations from which tactile readings or impact sounds are collected. Particle filters are used to estimate the posterior density of a latent variable given observations. Here, observations are either tactile sensor readings when touching the object or impact sounds excited at the contact locations. The latent variable is the current contact location on the object's surface. For touch, we extract features from an FCRN network~\cite{laina2016deeper} pre-trained for depth prediction from tactile images. For audio, we extract MFCC features from each 3s impact sound. We compare these features with particles sampled from the object surfaces that represent the candidate contact locations. Particles with high similarity scores to the features of the actual tactile sensor reading or impact sound are considered more likely to be the true contact location. In each iteration, we weight and re-sample the particles based on the similarity scores, and then update the particles' locations based on the relative translations between two consecutive contacts obtained from the robot end-effector. We choose the 10 particles with the highest similarity scores as the candidate contact locations. For each object, we iterate the above process for 5-7 times until the predicted current contact location converges to a single location on the object's surface. We perform experiments both in simulation and in real world.

\begin{figure}
    \center
    \includegraphics[scale=0.265]{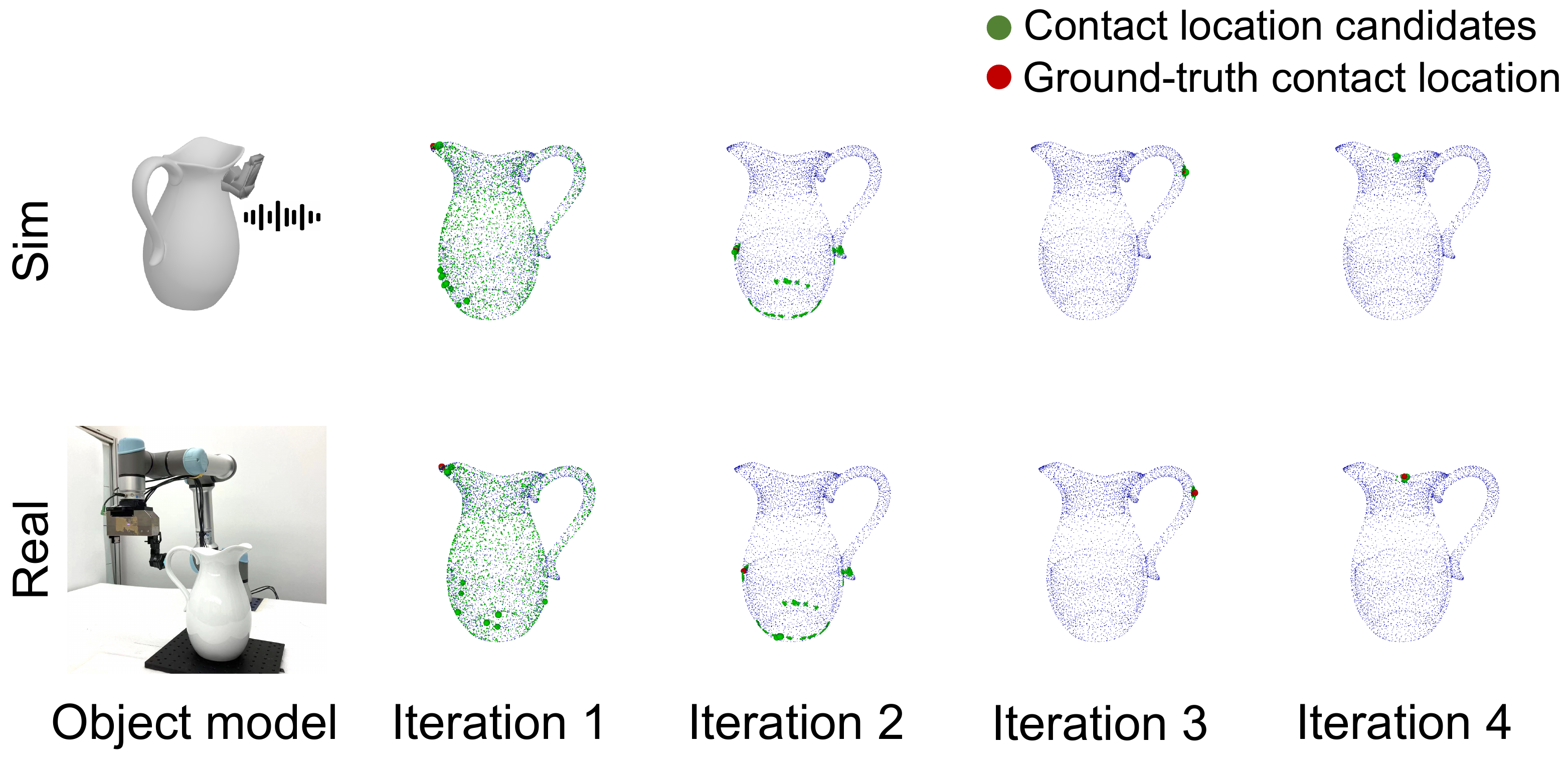}
    \caption{Qualitative results for contact localization with touch readings and impact sounds. Top: in simulation, bottom: real-world experiments. The candidate contact locations are shown as green particles in the particle filter. After several iterations shown from left to right in each row, the green particles converge to the ground-truth contact location shown as the red particle.}
    \label{fig:contact_localization_qualitative}
    \vspace{-0.1in}
\end{figure}

Table~\ref{Table:contact_localization} shows the results for six objects of complex shapes. We use the mean Euclidean distance with respect to the ground-truth contact location as the evaluation metric similar to~\cite{bauza2020tactile}. We compare the localization accuracy for using only touch readings, impact sounds, or their combinations, and a baseline that randomly predicts a surface position as the contact location. We can see that touch-based contact location is much more accurate than using audio. Combining the two modalities leads to the best Sim2Real performance. Fig.~\ref{fig:contact_localization_qualitative} shows a qualitative example for tactile-audio contact location with the pitcher object.

\begin{table*}[ht]
\small
\begin{tabular}{lcccccccccccc}
\toprule
\multirow{2}{*}{Modalities} & \multicolumn{2}{c}{\includegraphics[height=.05\textwidth]{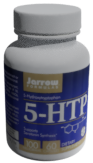}} &   \multicolumn{2}{c}{\includegraphics[height=.05\textwidth]{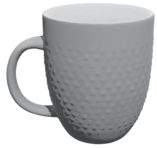}} & \multicolumn{2}{c}{\includegraphics[height=.05\textwidth]{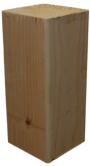}} & \multicolumn{2}{c}{\includegraphics[height=.05\textwidth]{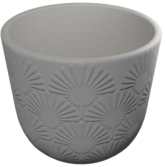}} &
\multicolumn{2}{c}{\includegraphics[height=.05\textwidth]{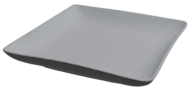}} &
\multicolumn{2}{c}{\includegraphics[height=.05\textwidth]{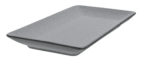}} 
\\ 
\cmidrule(lr){2-3}\cmidrule(lr){4-5} \cmidrule(lr){6-7} \cmidrule(lr){8-9} \cmidrule(lr){10-11} \cmidrule(lr){12-13} 
                  & Sim   & Real    & Sim   & Real & Sim & Real  & Sim   & Real    & Sim   & Real & Sim & Real  \\ 
\midrule
Average   &    2.12     &     2.01   &    2.97     &     1.91       &  4.80  &  3.26  & 4.53 & 4.49  & 2.44 & 2.53 &  2.52   &  3.29 \\ 
Vision   &    0.25     &     0.32     &    0.30     &     0.72      &  0.51   &  0.74  & 0.38 & 0.66 & 0.32 & 0.40 &  0.49   &  0.99 \\ 
Touch        &  0.24      &   0.56    &  0.29      &   0.80     &     0.35     &  0.61  &  0.38  & 0.43   &  0.30  & 0.41 &     0.36       &  1.11  \\
Vision + Touch        &  0.09      &   0.25    &  0.18      &  0.46     &     0.26       &  0.43  &  0.24  & 0.32   &  0.18  & 0.24 &     0.23       &  1.20  \\ \bottomrule
\end{tabular}
\caption{Results on visuo-tactile shape reconstruction. We report the Chamfer-L1 distance w.r.t. the ground-truth meshes in centimeters.}
\vspace{-0.1in}
\label{Table:shape_reconstruction}
\end{table*}

\subsection{Visuo-Tactile Shape Reconstruction}

Single-image shape reconstruction has been widely studied in the vision community~\cite{mescheder2019occupancy,choy20163d,park2019deepsdf,chang2015shapenet}. However, in cases where there is occlusion such as during dexterous manipulation, tactile signals become valuable for perceiving the shape of the objects. Vision provides coarse global context, while touch offers precise local geometry. Here, we train models to reconstruct the shape of 3D objects from a single RGB image containing the object and/or a sequence of tactile readings on the object's surface.

We use Point Completion Network (PCN)~\cite{yuan2018pcn}, a learning-based approach for shape completion, as a testbed for this task. For touch, we use 32 tactile readings and map the associated deformation maps to a sparse point cloud given the corresponding touching poses. The sparse point cloud is used as input to the PCN network for generating a dense and complete point cloud. For vision, instead of using a series of local contact maps as partial observations of the object, a global feature extracted from a ResNet-18 network from a single image containing the object is used to supervise the shape completion process. For shape reconstruction with vision and touch, we use a two-stream network that merges the predicted point clouds from both modalities with a fully-connected layer to predict the final dense point cloud. See Supp. for details.

\begin{figure}[t]
    \center
    \includegraphics[width=\linewidth]{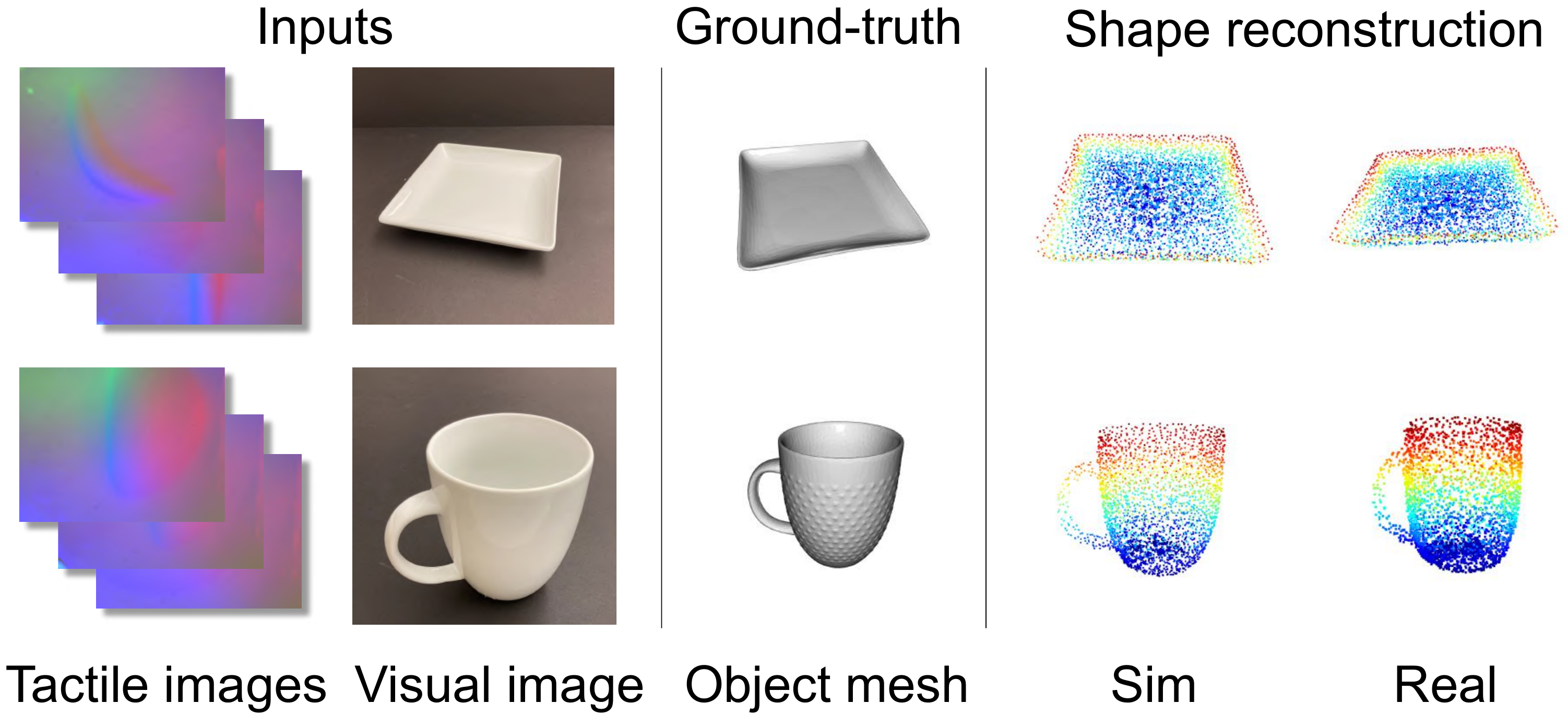}
    \vspace{-0.1in}
    \caption{Qualitative results for visual-tactile shape reconstruction in simulation (Sim) and real-world (Real) for the square tray and the coffee mug.}
    \label{fig:shape_reconstruction_qualitative}
    \vspace{-0.2in}
\end{figure}

Table~\ref{Table:shape_reconstruction} shows the results for six objects of different shapes. Compared to the ``Average" baseline that uses the average ground-truth mesh of the 6 objects as the prediction, shape reconstructions from a single image and a sequence of touch readings perform much better. Combining the geometric cues from both modalities usually leads to the best Sim2Real transfer performance. Fig.~\ref{fig:shape_reconstruction_qualitative} shows some qualitative results for shape reconstruction with vision and touch. We can see that the predicted point clouds in both simulation and real-world experiments accurately capture the shapes of the two objects, and matches the ground-truth object meshes well.

\section{Broader Impact and Limitations}

We will release our dataset and code upon publication of the paper, so that it can be easily accessible to the community as a standard benchmark for multisensory learning. This avoids the need to purchase real-world objects for such tasks, and can especially benefit people in areas where international shipping and purchasing of specific real-world objects is challenging. Furthermore, our implicit representation is computationally much cheaper to render multisensory data compared to the initial multisensory simulation, which is potentially more environmentally friendly. 

Bridging the gap between sim and real for multisensory object-centric learning is inherently difficult. While we have shown Sim2Real transfer for a series of objects, the objects in our dataset are all rigid-body objects, and we assume single homogeneous material for the whole object. However, real-world objects are complex and often contain several parts, which can be non-rigid and are of different material types. Furthermore, the 3D space in which these objects are located is of diverse lighting/noise conditions, reverberation effects, etc. Sim2Real object transfer is challenging without modeling all these factors, which we leave as future work.

\section{Conclusion}

\name~is a dataset of 1,000 objects in the form of implicit neural representations aimed at advancing multisensory learning in computer vision and robotics. Compared to existing work, our dataset is 10 times larger in scale and orders of magnitude faster in rendering time. We also significantly improve the quality and realism of the multisensory data. We show that models learned with our virtualized objects successfully transfer to their real-world counterparts on three challenging tasks. Our dataset offers a promising path for multisensory object-centric learning in computer vision and robotics, and we look forward to the research that will be enabled by~\name.

\begin{small}
\paragraph{Acknowledgements.} We thank Sudharshan Suresh, Mark Rau, Doug James, and Stephen Tian for helpful discussions. The work is in part supported by the Stanford Institute for Human-Centered AI (HAI), the Stanford Center for Integrated Facility Engineering, NSF CCRI \#2120095, Toyota Research Institute (TRI), Samsung, Autodesk, Amazon, Adobe, Google, and Facebook.
\end{small}


{\small
\bibliographystyle{ieee_fullname}
\bibliography{ref.bib}
}

\end{document}